\documentclass[journal,11pt,onecolumn,draftclsnofoot,]{article}

\usepackage{lipsum}
\usepackage{setspace}
\usepackage[version=4]{mhchem}
\usepackage{siunitx}
\usepackage{lscape}
\usepackage{authblk}
\usepackage[english]{babel}
\usepackage{graphicx}
\usepackage{booktabs}
\usepackage{blindtext}
\usepackage{hyperref}
\usepackage{xcolor}
\usepackage{threeparttable}
\usepackage[normalem]{ulem}
\usepackage{url}
\usepackage{caption}
\usepackage[margin=1in]{geometry}
\usepackage[most]{tcolorbox}

\usepackage{natbib}
\begin{document}

\begin{titlepage}
\title{Classification of animal sounds in a hyperdiverse rainforest using Convolutional Neural Networks with data augmentation \\[1ex] \large Running head: CNN for rainforest sound classification}

\author[1]{Yuren Sun}
\author[2,3]{Tatiana Midori Maeda}
\author[4,5]{Claudia Sol\'{i}s-Lemus}
\author[4,6*]{Daniel Pimentel-Alarc\'{o}n \thanks{corresponding author, pimentelalar@wisc.edu}}
\author[2,3*]{Zuzana Bu\v{r}ivalov\'a \thanks{corresponding author, burivalova@wisc.edu}}
\affil[1]{University of Wisconsin-Madison, Department of Computer Science}
\affil[2]{University of Wisconsin-Madison, The Nelson Institute for Environmental Studies}
\affil[3]{University of Wisconsin-Madison, Department of Forest and Wildlife Ecology}
\affil[4]{University of Wisconsin-Madison, Wisconsin Institute for Discovery}
\affil[5]{University of Wisconsin-Madison, Department of Plant Pathology}
\affil[6]{University of Wisconsin-Madison, Department of Biostatistics and Medical Informatics}

\maketitle
\end{titlepage}

\section*{Abstract}
{\parindent0pt

To protect tropical forest biodiversity, we need to be able to detect it reliably, cheaply, and at scale. Automated detection of sound producing animals from passively recorded soundscapes via machine-learning approaches is a promising technique towards this goal, but it is constrained by the necessity of large training data sets. Using soundscapes from a tropical forest in Borneo and a Convolutional Neural Network model (CNN), we investigate $i)$ the minimum viable training data set size for accurate prediction of call types ('sonotypes'), and $ii)$ the extent to which data augmentation and transfer learning can overcome the issue of small and imbalanced training data sets. We found that even relatively high sample sizes ($>80$ per sonotype) lead to mediocre accuracy, which however improved significantly with data augmentation and transfer learning, including at extremely small sample sizes ($3$ per sonotype), regardless of taxonomic group or call characteristics. Neither transfer learning nor data augmentation alone achieved high accuracy. Our results suggest that transfer learning and data augmentation could make the use of CNNs to classify species' vocalizations feasible even for small soundscape-based projects with many rare species. Retraining our open-source model requires only basic programming skills which makes it possible for individual conservation initiatives to match their local context, in order to enable more evidence-informed management of biodiversity

\providecommand{\keywords}[1]
{
  \small	
  \textbf{\textit{Keywords: }} #1
}
\keywords{Bioacoustics,  Convolutional Neural Network, Conservation, Data Augmentation, Passive Acoustic Monitoring, Sound Classification, Tropical forest,  Transfer Learning}
}

\section{Introduction}

\label{introSec}

The extinction of species presents an irreversible loss to humanity, and preventing biodiversity loss is one of the biggest challenges our society faces \citep{ceballos2010sixth}. Tropical forests are some of the most species-rich terrestrial ecosystems, yet they are highly threatened by human activities, such as deforestation, hunting, mining, and selective logging \citep{betts2017global}. An increasing number of conservation projects strive to be evidence-informed, which often requires being able to effectively monitor biodiversity over time and space \citep{burivalova2019works}. However, in tropical forests, the monitoring of animal biodiversity at an appropriate scale and level of replication is notoriously difficult through direct observational studies \citep{hillebrand2018biodiversity}, and the dense  tropical forest canopy does not allow for most fauna species to be seen on satellite images \citep{burivalova2019sound}. While remote sensing to track conservation outcomes such as deforestation is relatively advanced, there exist no broadly accepted methods to passively monitor biodiversity across time and space.

Biodiversity monitoring is undergoing a transformation through the synergy of new data streams, better data storage, and novel analytical techniques that can turn `big biodiversity data' into insights. Soundscapes, defined as the collections of all sounds in a given landscape \citep{sueur2015ecoacoustics}, typically obtained through passive acoustic monitoring, are a particularly important new data stream  in tropical forests. This is because many species, across several taxonomic groups (birds, invertebrates, mammals, amphibians) vocalize or make sounds to communicate  \citep{fletcher2014animal}. Soundscapes can be used either to gain insights about the entire biophony (all vocalizing species in the landscape), by calculating various soundscape indices \citep{buxton2018efficacy,sethi2020characterizing,bradfer2019guidelines}, or by identifying individual species through expert analysis, template matching or various machine learning methods, such as Convolutional Neural Networks (CNN) \citep{zhong2020multispecies,wood2019detecting}.

CNNs are becoming increasingly popular in the detection and classification of biodiversity based on sound, as they enable the processing of much larger datasets than experts could examine manually, and account for natural variation in sounds better than template matching techniques \citep{zhong2020multispecies,lebien2020pipeline,stowell2019automatic,ko2018convolutional}.  Existing studies typically focus on a few, common species, such as a pre-trained CNN to classify among 24 bird species in Puerto Rico, based on a training data set of 100,000 positive (presence of a call of the species of interest) and 243,000 negative (absence of species) data points \citep{zhong2020multispecies,lebien2020pipeline}. Perhaps the most successful recent CNN, BirdNet, focuses on the arguably best-known animal group on Earth - North American and European birds \citep{kahl2021birdnet}. BirdNet is able to classify hundreds of bird species, based on a training dataset with thousands of data points even for less common species \citep{kahl2021birdnet}.

These highly successful CNN applications present unprecedented advances for biodiversity monitoring. Yet, these methods may not be directly applicable in cases where \textit{all} vocalizing species are of interest, such as in rapid biodiversity surveys, inventories, or prioritization projects. In such cases, metrics akin to species richness and community composition are needed, and rare species, naturally occurring at low densities or in limited areas, might be especially important to detect. Being able to detect all `sonotypes' (sound types, assumed to be correlated with species richness) would enable estimating such measures of biodiversity, even if not all sonotypes can be identified to a species level.

There are  challenges to sound-based recognition of individual species in any natural soundscape, such as due to a variable distance of the sound source (the animal) to the sensor (microphone); inter-species and inter-individual variation in vocalizations; multiple unique vocalizations (sonotypes) per species, including mimicry; or biases due to equipment \citep{towsey2012toolbox,darras2020high}. Additional factors make such species recognition exponentially harder in tropical forests, which are often extremely diverse in terms of vocalizing species, some of which are as yet unknown. This hyper-diversity can result in partial overlap in individual species' vocalizations, such as between continuous cicada choruses and bird vocalizations \citep{aide2017species}. Dense vegetation results in signals that attenuate quickly and in a non-uniform way \citep{rappaport2020acoustic,darras2016measuring}. An important hurdle in tropical forests is the heavily skewed distribution of species: whereas there are a few common ones, there is typically a large number of rare species, making it challenging to create balanced training data sets \citep{yang2021systematic}. Models based on imbalanced data set are more likely to over-fit to common species at the expense of rare ones, which are  most likely to be of conservation concern \citep{schneider2020three}.

Every level of variation in a species' vocalization, be it due to the species' behaviour or the physical environment, requires additional training data for a machine learning model to be successful. Indeed, generating sufficient training data sets is a bottleneck in uptake of this technology in conservation \citep{lamba2019deep}, especially for small to medium-size conservation projects, which often rely on undergraduate and graduate students from local universities to implement scientific projects \citep{johnson2022more}. We set out to address this  limitation by creating a model accessible to scientists with basic programming skills, by investigating the minimum necessary training data size, and by testing approaches that can boost performance at small and imbalanced sample size.

Specifically, we investigate the feasibility of using transfer learning to create CNN models to classify \textit{all} audible sounds emitted by birds, mammals, amphibians and invertebrates. Transfer learning is the process of importing insights learnt in related tasks. This is done by initializing a neural network’s parameters with those obtained after training another network on a similar task(section \ref{transfer-learning}). We test our models on a set of exhaustively, manually labeled and segmented soundscapes from a hyperdiverse rainforest in Indonesian Borneo.  Our aim is to $i)$ identify the minimum viable size of a balanced and imbalanced training data set that would allow a reasonable classification accuracy; and $ii)$ test whether we can achieve an improved classification performance with data augmentation and transfer learning. Data augmentation is a technique that artificially increases the sample size of the training dataset, such as through slightly distorting, rotating, or shifting the original data (section \ref{data-augmentation}). Our overarching goal is to create a robust, open-source model to classify rare as well as common sound types from rainforest soundscapes, such that our model could be easily re-trained for new landscapes, integrated with automated segmentation techniques, and ultimately adapted and reused by individual conservation projects in need of fauna monitoring.

\section{Materials and Methods}

\subsection{Study Site and Soundscape Data}
The data set that we used is publicly available on the bioacoustic workbench Ecosounds (www.ecosounds.org). It consists of a selection of soundscape recordings collected at 15 sites in the tropical rainforests of Berau and East Kutai Regencies in East Kalimantan, Indonesia, from June 2018 to June 2019, within a selective logging concession \citep{burivalova2019sound}. The soundscapes were recorded with autonomous, mono Bioacoustic Recorders (Frontiers Lab), at 2 m above ground, pre-programmed to record continuously in 30-min segments, at 40 dB gain, and at a 44.1 kHz sampling rate. The devices were programmed at variable schedules throughout the year, and at sites that were at least 600 m away from each other to ensure non-overlapping soundscapes; at least 200 m distant from the nearest active or inactive logging roads, ridgeline foot trails, or rivers to avoid potential human interference with the devices; and at altitudes ranging from 387 to 517 m. For our experiment, we selected minutes for exhaustive manual annotation as follows. First, at two sites, we selected one minute at random within a 1 hour period in the morning (dawn $\pm{30}$ min) and evening (dusk $\pm{30}$ min), and sampled the same minute once a month for 12 months, in order to capture seasonal diversity. Then, we selected  the same dawn minute from 15 sites, but only during one day, in order to capture spatial diversity. This resulted in a total of 63 sample minutes, encompassing the period of the highest acoustic diversity, across both time and space. Each soundscape captures a mix of geophony (rain, thunder, wind), anthropophony (airplanes, machinery), and biophony (all animal vocalizations) \citep{sueur2015ecoacoustics}.

Using recordings and spectrograms in  Raven Pro 1.5 \citep{Raven}, an expert sound analyst manually annotated 3629 animal vocalizations $V_{1},V_{2},\dots,V_{3629}$ belonging to 448 sonotypes. A sonotype is ``a note or series of notes that constitute a unique acoustic signal''\citep{aide2017species}. The analyst also assigned each sonotype into one of four higher taxonomic groups: birds, invertebrates, mammals, and amphibians. For example, if the first two sounds encountered ($V_{1},V_{2}$) were the same bird call, we labelled them both `sonotype 1', and `bird'. If the third encountered sound $V_{3}$ was an insect, we labelled it `sonotype  2' and `insect'. The same sound would be labelled as the same sonotype regardless which site and during which time it was encountered (Fig. \ref{FigS1}). Vocalizations spaced by more than 2s were labeled separately, even if they were of the same sonotype (Fig. \ref{FigS1}). It is important to note that one species can make multiple vocalizations, such as a song and call in the case of birds. These would be labelled as two separate bird sonotypes. Therefore, the number of sonotypes does not equal the number of species. All data were examined by the same analyst (TMM), who had previous experience with acoustic fauna identification. When the analyst was uncertain about a sonotype’s taxonomic group, three other specialists were consulted. Sonotypes that could not be assigned to a category even after specialist consultation were classified as unknown.

\begin{figure}[ht]
\centering
\includegraphics[width=16cm]{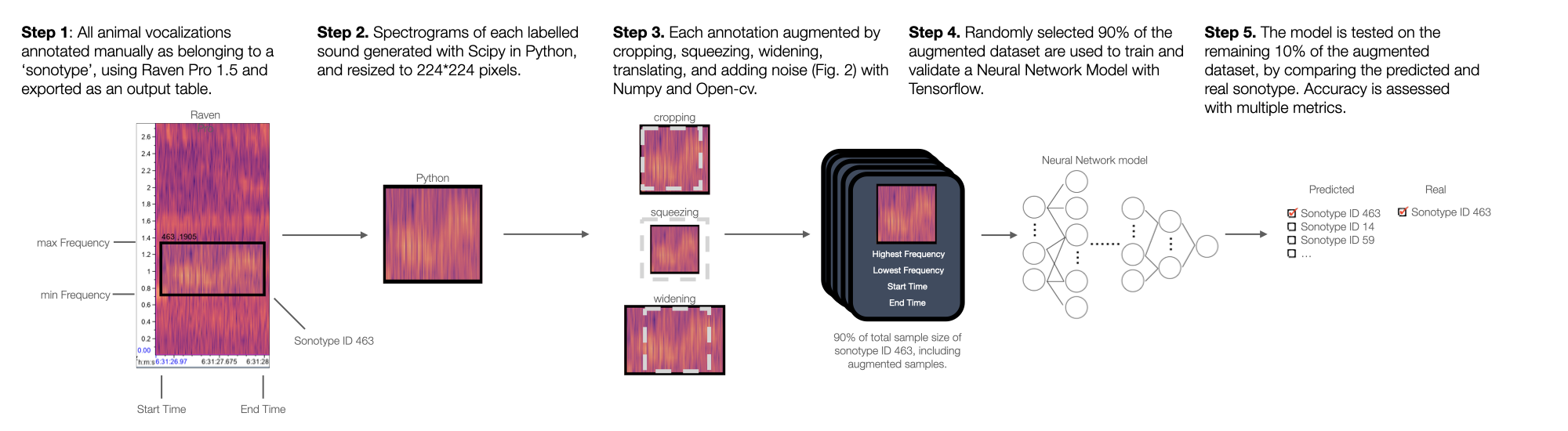}
\caption{Illustration of the sonotype classification work flow: Each sound is delimited with a box and labeled as a sonotype by an expert analyst (Fig. \ref{FigS1}). Then, it is augmented to 16 samples by means of several data augmentation techniques (Fig. \ref{augmented_sonotypes}).The augmented samples, along with their time and frequency ranges, become input for the neural network model (Fig. \ref{model_png}). The model has the goal to assign the sonotype ID to a new sound sample, reserved from the original data as a testing data set.}
\label{process}
\end{figure}

We processed each soundscape sample to obtain its spectrogram, using Scipy \citep{2020SciPy-NMeth}. We used the default settings of scipy.signal.spectrogram with Tukey window of size 256, shape parameter equal to 0.25, and number of points to overlap between segments/windows of 32. The samples were encoded in a $129 \times 353,389$ image depicting the frequency components of the soundscape in the 0 to 22,050 Hz range. The height ($129$) is derived from 256 window size, with one more index for frequency 0. The width $353,389$ represents the length of the time array/sequence, and is determined by how many segments/windows we can achieve from the input audio. In other words, it is derived from the input audio length, window size, and overlap. The audio files that were read from 30-minute wav files have length $79,159,274$. With the default scipy.signal.spectrogram function \citep{2020SciPy-NMeth}, the width is computed by (length of audio file - overlap) / (window size - overlap), which is $(79,159,274 - 32) / (256 - 32) = 353,389.473$ and is rounded to $353,389$. Each vocalization $V_i$ was encoded in a gray-scale image, enclosing the portion of the spectrogram from the initial to the final times of the vocalization, and from the lowest to the highest dominant frequencies of the vocalization, keeping the information on time and frequency bounds for later use (see Fig. \ref{process}). The slight information loss due to this encoding was necessary to use the ImageNet model for transfer learning (see below). Given a new vocalization $V,$ our goal was to automatically determine its sonotype. To this end, we propose the deep neural network architecture detailed below.

\subsection{Transfer Learning and Pre-processing}
\label{transfer-learning}

Deep learning classifiers generally rely on the availability of a large number of samples for each class (i.e. sonotype). Successful data sets range in the order of thousands to millions of samples per class. Due to the high level of expertise and time required to identify and classify vocalizations into sonotypes, our data set is orders of magnitude smaller (Fig. \ref{histogram}A). Yet, it is representative of what a small to mid-size conservation project may be able to achieve in terms of manually labeling examples, typically in collaboration with a paraecologist from a local community or an undergraduate student \citep{johnson2022more}. Deep learning models tend to overfit small data sets, producing a large generalization error. To address this issue, the machine learning community has developed {\em transfer learning} and {\em data augmentation}.

Transfer learning is the process of importing insights learnt in related tasks. This is done by initializing a neural network's parameters with those obtained after training another network on a similar task. For example, one can initialize a neural network that aims to learn Spanish with the parameters of a network that already knows French. Transfer learning has been shown to improve learning speed and accuracy with fewer data, reduce generalization error and improve the overall network performance  \citep{yosinski2014transferable}. In recent years, highly successful image classification architectures have been produced, such as  VGG-19 \citep{chollet2015keras} and ResNet-50 \citep{he2016deep}, and well-organized image databases that can be used to pre-train models, such as  ImageNet \citep{ILSVRC15}.

We transferred the parameters of the Keras VGG-19 model \citep{chollet2015keras} after being trained on the ImageNet data set \citep{ILSVRC15}. Having considered different options, we deemed ImageNet as the most appropriate since we are classifying soundscapes images (spectrograms), and since ImageNet is now a standard, well-studied and accessible transfer learning tool. The first step towards this endeavor was to pre-process our data to match the Keras VGG-19 format, which is 8-bit per RGB (red, green, blue) channel, $224 \times 224$ colored images. To this end, we first adjusted all vocalizations to a common scale, using Numpy's  linear normalization \citep{2020NumPy-Array}. More precisely, we identified the highest and lowest frequency intensities $F_i$, $f_i$ of each vocalization image $V_i$, and performed the following transformation, to obtain the normalized vocalization $V_i'$:

\begin{align}
\label{normalization}
V_i' \ = \ \frac{V_i \ -\ f_i}{F_i \ - \ f_i} \ \cdot \ 255
\end{align}

After this procedure, each pixel in the normalized vocalization image $V_i'$ will take a value between 0 and 255, corresponding to the standard 8-bits format used to store single-channel gray-scale images. To transform $V'_i$ to the required colored image, we simply replicated each $V_i'$ into the three RGB channels, and resized each image to $ 224 \times 224$ pixels with Open-cv \citep{opencv_library}. All data (including vocalization images, time and frequency ranges, and classification labels) were stored in hdf5 format using h5py \citep{hdf5}, as is standard practice for large data sets.

\subsection{Data Augmentation}
\label{data-augmentation}

Data augmentation aims to artificially increase the sample size. Common augmentation methods involve image flipping, shifting, zooming in or out, rotating, and distorting. For instance, a study on image classification with such augmentation methods found an increase in accuracy by 7 percentage points  when classifying cats and dogs and by 4 percentage points when classifying dogs and goldfish \citep{wang2017effectiveness}. Other studies have augmented data sets by transposing, squeezing, or stretching the images \citep{nanni2020data}; or augmented data sets with regional dropout and mixing images, in a technique called cutmix \citep{yun2019cutmix}. For speech recognition, another technique, called specAugment, augments data sets by time warping and frequency and time masking \citep{park2019specaugment}.

\begin{figure}[ht]
\centering
\includegraphics[width=9cm]{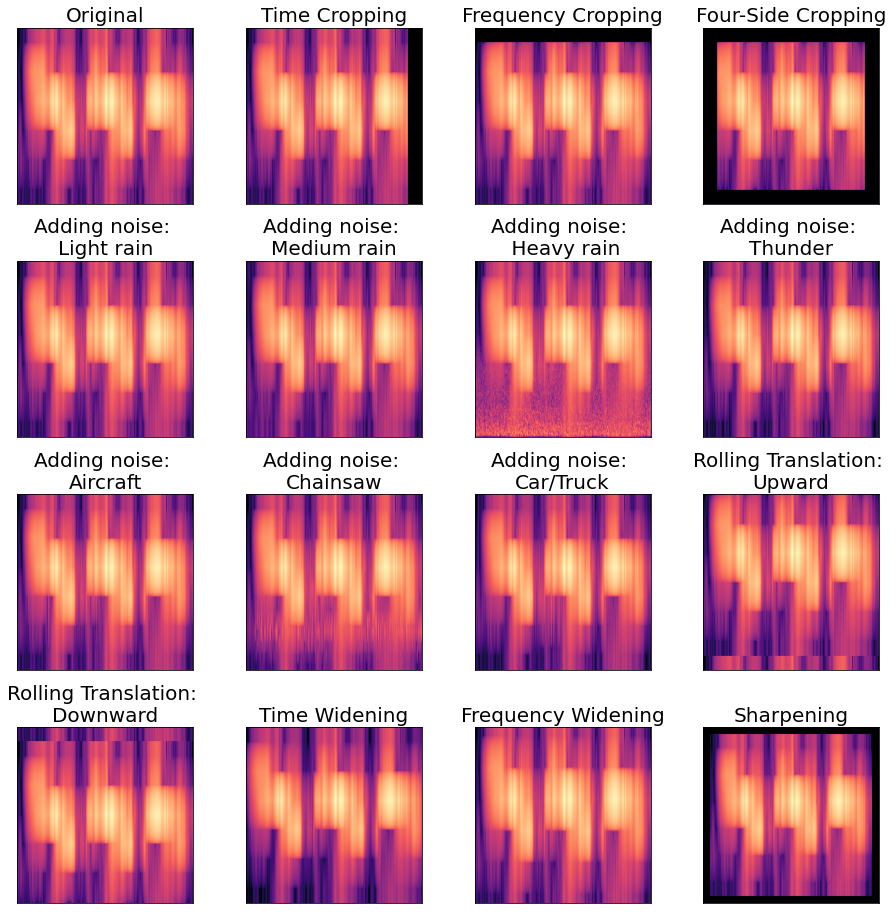}
\caption{Illustration of the different techniques that we used to augment our data set of labeled rainforest sounds. Depicted are spectrograms, with time on the x axes and frequency on the y axes. Brighter colours represent higher amplitude.}
\label{augmented_sonotypes}
\end{figure}

We selected data augmentation techniques (Fig. \ref{augmented_sonotypes}) to produce variation in our spectrograms while maintaining a realistic vocalization pattern: \textit{i)}
Cropping of the spectrogram's time range (X axis of the spectrogram), frequency range (Y axis), or both; \textit{ii)} adding the sound of light, medium or heavy rain, thunder, aircraft, chainsaw, and car/truck to the spectrogram; \textit{iii)} translating the whole spectrogram up or down in terms of frequency; \textit{iv)} widening the spectrogram in the time or frequency range; \textit{v)} sharpening the spectrogram by squeezing it in time and frequency range.

We implemented the augmentation methods with Numpy \citep{2020NumPy-Array} and Open-cv \citep{opencv_library}. When applying data augmentation with cropping, squeezing, widening, and translating, we modified the spectrogram by a random number between 5\% and 10\% of the size of the original spectrogram, as this reflects the approximate range of variation in nature. When adding noise, we first normalized the noise in the same way as the original spectrogram (see eq. \eqref{normalization}). Then, we added 1/3 of the noise, including the sound of  rain, thunder, aircraft, chainsaw, and car to the original spectrogram, to reflect a typical amount of background noise and normalized the new spectrogram again.

\subsection{The Deep Learning Model}
\label{The neural network model}

\begin{figure}[ht]
\centering
\includegraphics[width=12cm]{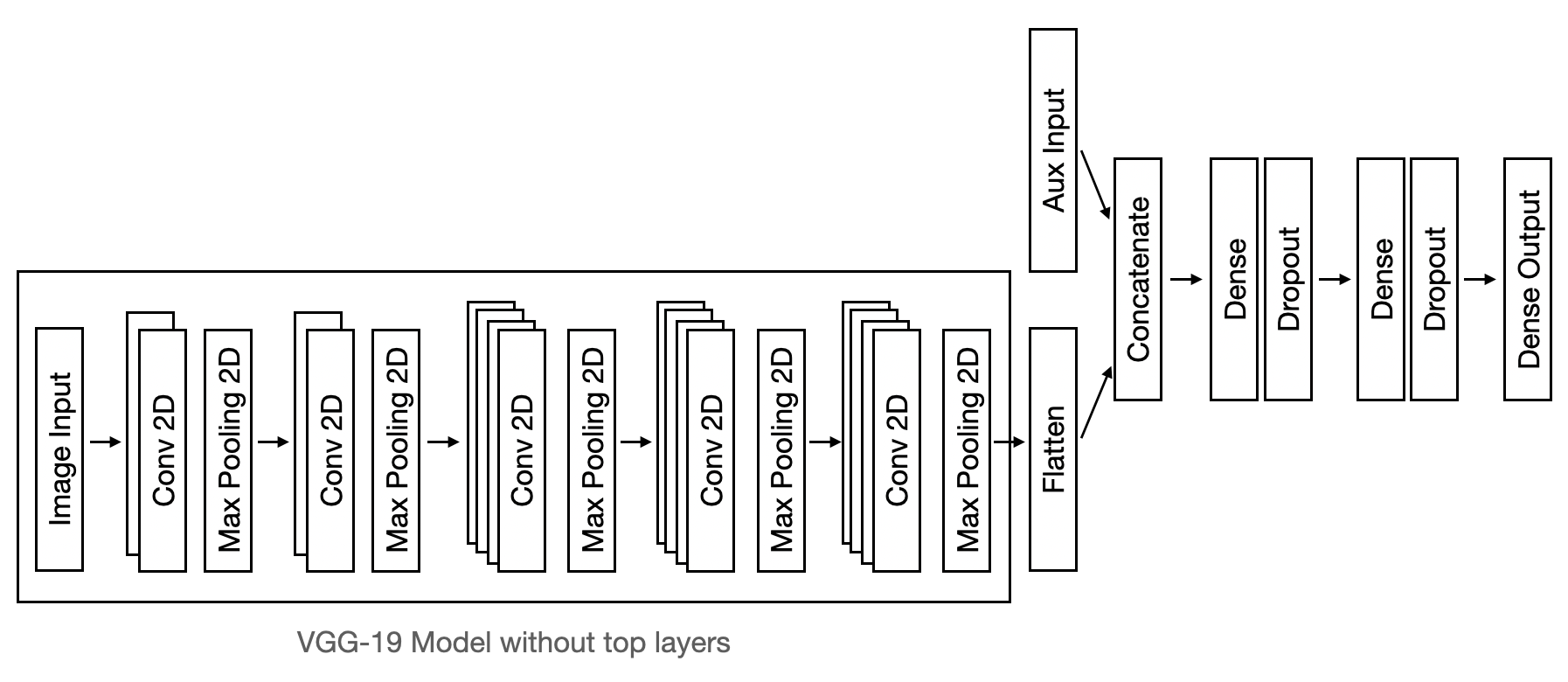}
\caption{Architecture of our Deep Learning model. The Image Input are spectrograms of animal sounds, and the auxiliary inputs (Aux Input) are the highest and lowest frequencies as well as the start and end times.}
\label{model_png}
\end{figure}

Our proposed neural network architecture passes the vocalization images $V_i$ through the architecture of the Keras VGG-19 model \citep{chollet2015keras} without the four top layers (Fig. \ref{model_png}). The output (image) of the Keras VGG-19 model is then flattened into a vector and concatenated with the four values corresponding to the start and end time, lowest and highest frequency of the vocalization (depicted as Auxiliary Input in Fig. \ref{model_png}). This vector is then passed through two couples of dense-dropout layers and a softmax dense output layer to obtain the final output.

\subsection{The Training Process}

The training was processed using TensorFlow \citep{tensorflow2015-whitepaper} and Keras \citep{chollet2015keras}. In the experiments where transfer learning was applied, we initialized the layers of the Keras VGG-19 model with weights trained on the Image-Net and froze these layers during training. When we trained the model without transfer learning, we initialized the model with random weights for each layer and did not freeze any layers during the training process. Overfitting and generalization errors would appear if we trained the model for too many epochs (iterations of the whole training data set). To avoid this issue, we applied early stopping to terminate the model appropriately. We decided whether to stop training based on validation loss, which represents how far the predicted outputs deviate from the expected output and whether the model is able to generalize well on data. The loss is computed with Keras categorical cross-entropy loss function for each epoch. If the model keeps generalizing to the data instead of overfitting, the validation loss will keep decreasing. We terminated the training process when the validation loss did not decrease for 15 continuous epochs, in other words, we applied early stopping with a patience of 15 epochs. We also applied checking points to the model that saved the model with the lowest validation loss, which represents the highest possibility to generalize well on the data, and used the model with the lowest validation loss as the final model for testing.

\begin{figure}[ht]
\centering
\includegraphics[width=13cm]{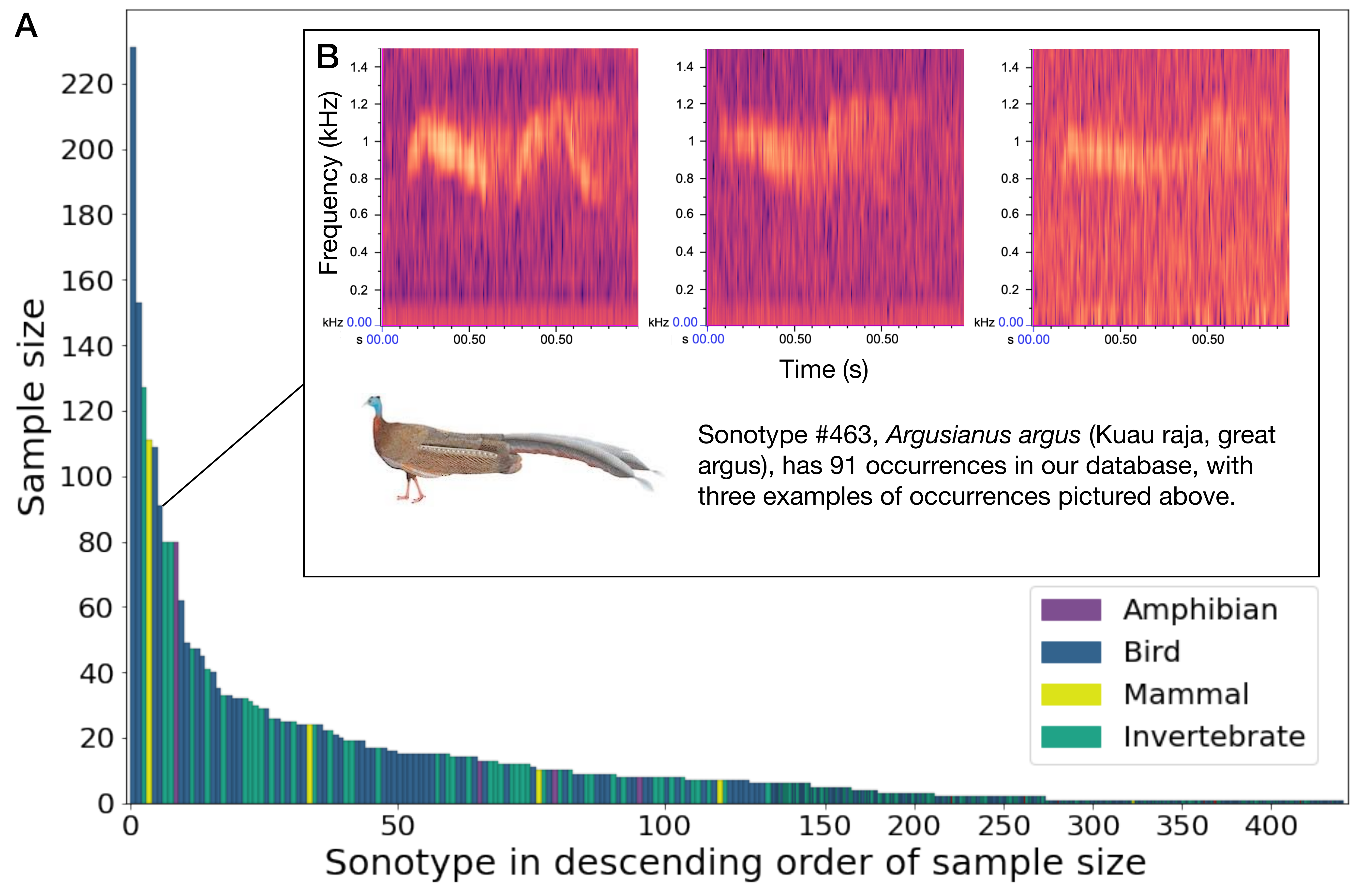}
\caption{A - Sonotypes (unique animal sound types) from 63 exhaustively labeled minutes of rainforest soundscapes, classified into 4 taxonomic groups, and ordered in descending sample size (number of occurrences of each sonotype). B - Spectrograms of three example occurrences of one sonotype, call of the Kuau raja (\textit{Argusianus argus}, illustration from \citep{HBW}).}
\label{histogram}
\end{figure}

\subsection{Experiments}

In each trial of all  experiments, we first split our available data (before augmentation) into training, validating, and testing sets. We used the training set to train the model and update its parameters; the validation set to evaluate whether we will continue or stop the training process; and the testing set to evaluate the model after training. To avoid contaminating our results with peeking bias (double-dipping), we discarded all sonotypes with fewer than 3 samples; from the total of 448 sonotypes, 212 had at least 3 samples. This guaranteed that each sonotype (class) had at least one independent sample for each purpose (training, validating, and testing). All sonotypes with 3 or more samples were respectively split to 80\%, 10\% and 10\% for training, validation, and testing, or in the closest possible proportion, so long as at least one independent sample was available for each purpose. 

We measured the performance of our model with accuracy, AUC, precision, recall, specificity, and f1 scores. The accuracy of our model is the fraction of correctly classified previously unseen vocalizations from the testing data set. AUC is the area under the receiver operating characteristic curve, and we present results as multi-class AUC, which is the average of AUCs for each sonotype. Precision is measured as $TP/(TP + FP)$, where TP is true positive, TN true negative, FP false positive, and FN false negative. We obtain class-wise mean average precision (cmAP), which is calculated by first computing the average precision, AP, for each sonotype, and then computing the mean of the APs across sonotypes without weighting on sample sizes. We also obtain mean average precision (mAP), which is calculated in a similar way to cmAP, except we weight on sample sizes when computing the mean of the APs across sonotypes. Recall is calculated as $TP/ (TP + FN)$, and specificity  as $TN/(TN + FP)$. Finally, f1 score is measured as $2 * (precision * recall) / (precision + recall)$. We average recall, specificity, and f1 score across sonotypes in each run and then average them across all the replicated runs, respectively.

In our first experiment, we studied the effect of data augmentation and transfer learning, while classifying amongst an increasing number of sonotypes. In each of the 95 independent sets of trials, we fixed the number of sonotypes, $K$, and varied $K$ from 2 to 6. Then we selected $K$ different sonotypes uniformly at random (among those that have $s\geq{49}$ samples), selected $s=49$ samples uniformly at random for each sonotype, and evaluated the performance of our model after training it with and without data augmentation and transfer learning. We fixed the sample size of each sonotype to 49 because there are only 11 sonotypes with at least 49 samples in our data set to achieve enough replicates. During data augmentation, we augmented each sample to 16 samples using all of the transformations described above (Fig. \ref{augmented_sonotypes}).

In our second experiment,  we investigated the effects of sample size on a balanced data set, with transfer learning and with and without data augmentation. In 35 independent sets of trials, we fixed the sample size $s$ for each sonotype and varied $s$ from 3 to 80, selecting $K=6$ different sonotypes uniformly at random among the ones that have at least $s$ samples. Then we selected $s$ samples from each of $K=6$ different sonotypes uniformly at random and evaluated the performance of our model after training it with and without data augmentation. For the former, we augmented data such that the 80-10-10 proportions of training, validating, and  testing data were maintained, selecting randomly from the  transformations described above, yielding a total of 200-25-25 samples per sonotype. As the largest sample size across all sonotypes is 231, augmenting the samples to a total of 250 samples per sonotype ensured an appropriate level of augmentation. We set the maximum sample size $s$ to 80 as there are only 6 sonotypes with at least 80 samples. Due to findings from the first experiment, we no longer tested the model without transfer learning in this and the following experiment.

To investigate the influence of an imbalanced data set - a principal challenge in training neural networks with a small data set -  we conducted a third experiment, similar to the previous one, but without fixing the sample size of sonotypes. Instead, we selected 6 sonotypes at random, and trained and tested the model with and without data augmentation, and with transfer learning. For each selection of 6 sonotypes, we calculated the  average and minimum sample size. In each of the 3114 independent trials, we selected $K=6$ different sonotypes at random to achieve a uniform distribution of average sample sizes and and performed the same experiment as before.

Finally, we analyzed whether classification performance is influenced by the taxonomic group, or other sonotype properties, such as minimum and maximum frequency, frequency range, and the duration of the sound. We tested the influence of these variables on the accuracy by doing an analysis of variance, with sonotype properties as explanatory variables.

\section{Results}

From the 63 sample minutes that we exhaustively labeled for all biophony, we obtained 3629 sounds emitted by fauna, falling under 448 sonotypes, within four broad taxonomic groups: birds, amphibians, invertebrates, mammals. Eight sounds were labeled as unknown, and additional 154 sounds were labeled as anthropophony or geophony, and used in data augmentation. Beyond a few common sonotypes, the vast majority  were rare (Fig. \ref{histogram}A). For example, there were only 
ten sonotypes with sample size $>50$; $76$ sonotypes with sample size $>10$; and $212$ sonotypes with sample size $\geq{3}$, which was the minimal sample size for a sonotype to be included in the following analyses. To produce a classification system for this highly imbalanced sonotype collection, we created an open-source CNN model (https://github.com/solislemuslab/tropical-stethoscope), pre-trained on ImageNet \citep{ILSVRC15}, with the Keras VGG-19 neural network architecture \citep{chollet2015keras}. We discuss the performance of our model mostly in terms of accuracy, as we found it to be correlated with other performance metrics (Tables \ref{tab-our-results},\ref{tab-aug-transfer-compare},\ref{tab-taxonomic-group}).

\subsection{Impact of data augmentation and transfer learning}

\begin{figure}[ht]
\centering
\includegraphics[width=15cm]{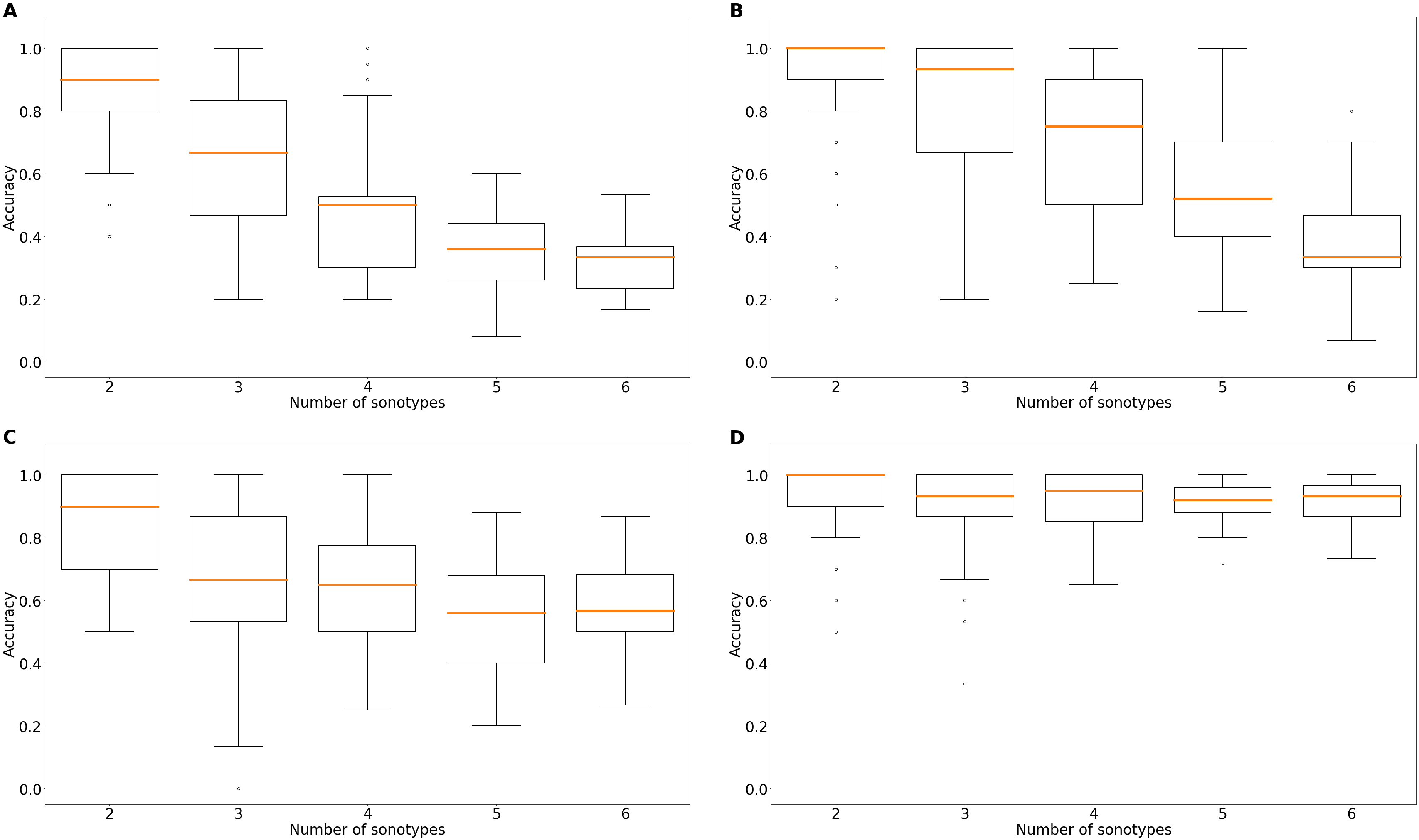}
\caption{Classification accuracy without data augmentation and without transfer learning (A), with data augmentation and without transfer learning (B), without data augmentation and with transfer learning (C), and with both data augmentation and transfer learning (D) among 2 - 6 sonotypes. See  Table \ref{tab-aug-transfer-compare} for other accuracy measures.}
\label{fix size under varied sonotype number}
\end{figure}

Without data augmentation and transfer learning, the model performance was poor, but better than a random guess (Fig. \ref{fix size under varied sonotype number}A). Data augmentation alone increased the average accuracy by 5.5 percentage points when classifying 2 sonotypes and by 6.7 percentage points when classifying 6 sonotypes (Fig. \ref{fix size under varied sonotype number}B). Transfer learning alone decreased the classification accuracy by 0.7 percentage points when classifying 2 sonotypes and increased the classification accuracy by 26.7 percentage points when classifying 6 sonotypes (Fig. \ref{fix size under varied sonotype number}C). Transfer learning together with data augmentation increased the average accuracy and by 7.9 percentage points when classifying 2 sonotypes and by 60.4 percentage points when classifying 6 sonotypes(Fig. \ref{fix size under varied sonotype number}D).  Both augmentation and transfer learning also reduced the variation in classification accuracy.

The benefits of both techniques were especially evident when increasing the number of sonotypes to classify. We found a decrease in classification accuracy with the increasing number of sonotypes (experiment 1, Fig. \ref{fix size under varied sonotype number}), but this  decrease was low when we used both data augmentation and transfer learning (Fig. \ref{fix size under varied sonotype number}D). The mean classification accuracy decreased by 54.7, 53.5, 27.3, and 2.2 percentage points when increasing the number of sonotypes from 2 to 6, when training the model without data augmentation without transfer learning (Fig. \ref{fix size under varied sonotype number}A), with data augmentation without transfer learning (Fig. \ref{fix size under varied sonotype number}B), without data augmentation with transfer learning (Fig. \ref{fix size under varied sonotype number}C), and with both data augmentation and transfer learning (Fig. \ref{fix size under varied sonotype number}D), respectively.

\subsection{Impact of sample size}

\begin{figure}[ht]
\centering
\includegraphics[width=17cm]{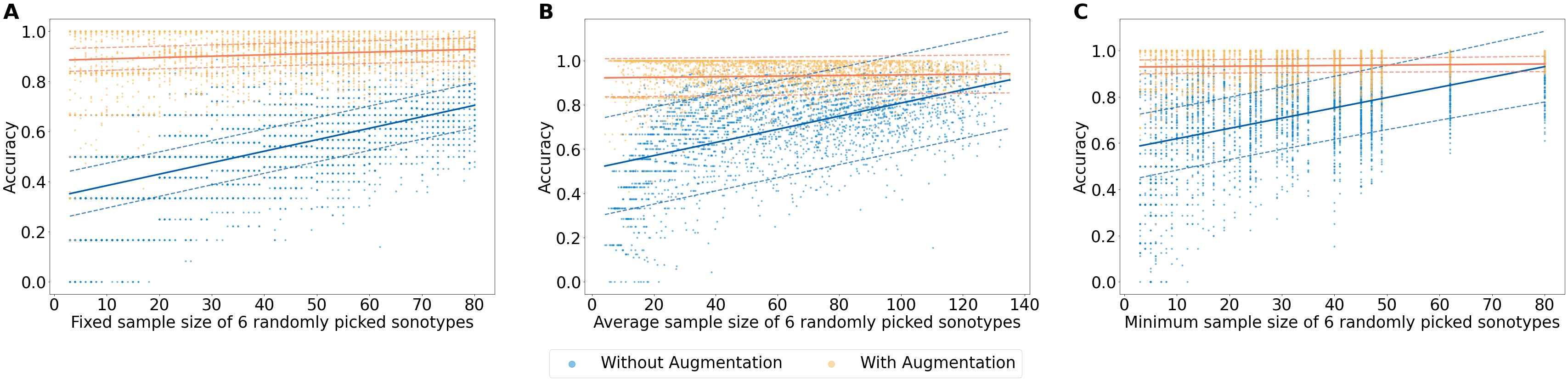}
\caption{Classification accuracy without (blue) and with (orange) data augmentation among 6 sonotypes. Each point represents one run of training, testing, and validation. (A) In each run, the sample size per sonotype was balanced, i.e., all six sonotypes were represented by exactly \textit{n} samples. (B) and (C) In each run, the sample size per sonotype was not set to be balanced. Accuracy is plotted against the average sample size (B) and minimum sample size (C). Lines represent the fits of linear regression, with 95\% CI. Random sonotypes were chosen every time.}
\label{6 random all}
\end{figure}

In a balanced dataset (experiment 2), we found an increase in accuracy with sample size (Fig. \ref{6 random all}A), and this effect was larger without data augmentation than with data augmentation (linear regression without augmentation: $p<0.01$, slope = 0.46, 95\% CI 0.41-0.50; with augmentation: $p<0.01$, slope = 0.05, 95\% CI 0.03-0.08). Data augmentation increased the classification accuracy by 39.0 percentage points on average (from $51.4\% \pm{18.2\%}$ to $90.4\% \pm{9.7\%}$) (Fig. \ref{6 random all}A, Table \ref{tab-our-results}). It also reduced the variability in accuracy by 8.5 percentage points, signifying a more stable classification performance. Data augmentation increased the AUC by 15.0 percentage points, from $82.9\%  \pm{12.7\%}$ to $97.9\% \pm{4.0\%}$ (Table \ref{tab-our-results}).

 In an imbalanced dataset (experiment 3), accuracy also increased with \textit{mean} and \textit{minimum}  sample size per sonotype (Fig. \ref{6 random all}B,C), and this effect was more pronounced without data augmentation, similarly to the balanced dataset. Data augmentation  increased  classification accuracy from $70.4\% \pm{17.7\%}$ to $93.2\% \pm{7.2\%}$, and the impact of data augmentation was even more pronounced on recall and f1 scores (Table \ref{tab-our-results}).

\begin{footnotesize}
\begin{table}[ht]

\caption{Accuracy measures for balanced and imbalanced data set without and with data augmentation, averaged across all tested sample size. mAP = mean average precision; cmAP = class-wide mean average precision; AUC = area under the receiver operating characteristic curve.}
\label{tab-our-results}

\begin{center}

\begin{tabular}{lrrrr}

 \toprule
& \multicolumn{2}{c}{Balanced} & \multicolumn{2}{c}{Imbalanced}\\
\cmidrule(l){2-3} \cmidrule(l){4-5} 
Augmentation	&	Without 	&	With  &	Without 	&	With \\

  \midrule
Accuracy	&   51.4	&   90.4	&   70.4	&   93.2		\\
mAP	&   57.5	&   90.4	&   75.2	&   93.2		\\
cmAP	&   57.4	&   90.4	&   73.4	&   93.2		\\
AUC	&   82.9	&   97.9	&   86.3	&   98.7		\\
Recall	&   51.4	&   90.4	&   60.4	&   93.2	\\
Specificity	&   81.6	&   97.0	&   84.7	&   98.1	\\
f1 score	&   42.8	&   89.2	&   55.3	&   92.6	\\

\bottomrule
\end{tabular}
\end{center}
\end{table}
\end{footnotesize}

\subsection{Factors influencing accuracy}

The taxonomic group which a particular sonotype belonged to did not influence the model performance, regardless of data augmentation, and  the classification accuracy was not influenced by the average or minimum frequency, or by the frequency range, indicating similar performance regardless of the sound-producing organism  (Fig. \ref{taxonomic group}).


\begin{figure}[ht]
\centering
\includegraphics[width=14cm]{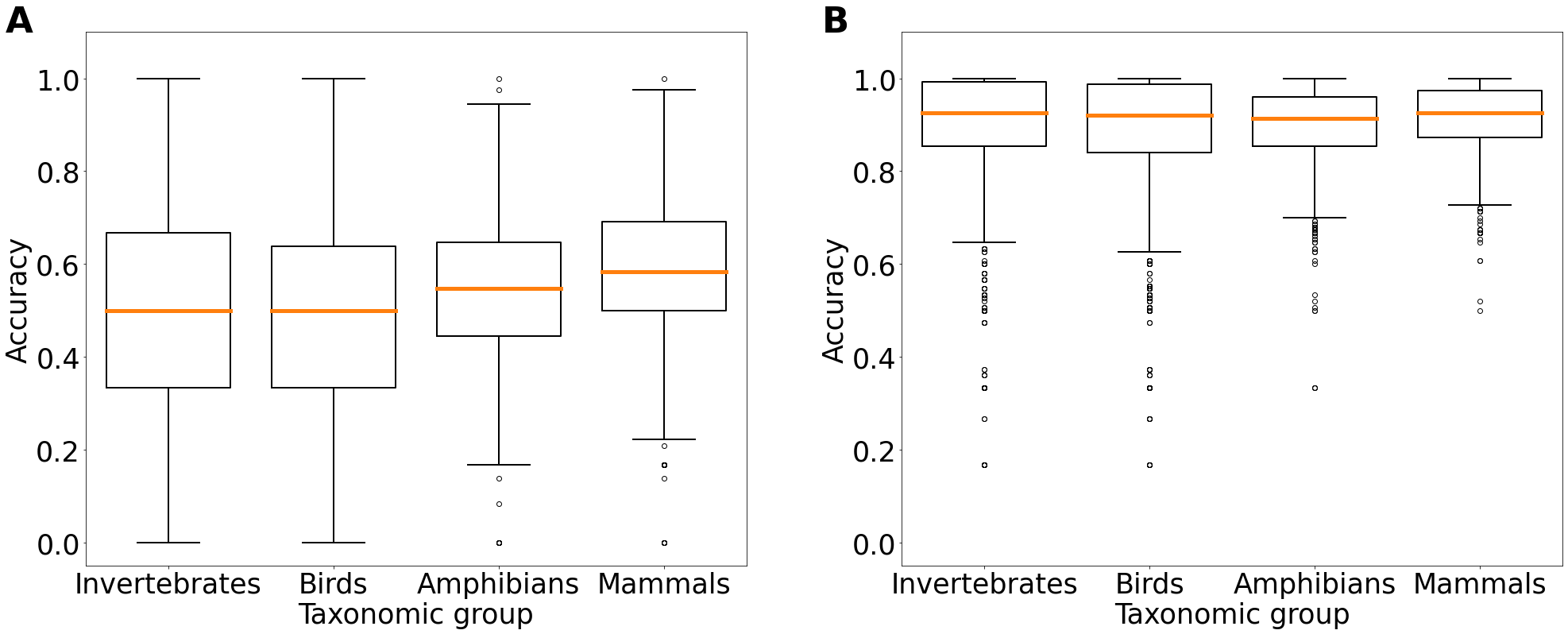}
\caption{Classification accuracy and taxonomic groups  without (A) and with (B) data augmentation. See table \ref{tab-taxonomic-group} for additional accuracy measures.
}
\label{taxonomic group}
\end{figure}

\section{Discussion}


Using a pre-trained CNN and a fairly limited training data set ($n=3629$) of high diversity (448 sonotypes), we were able to create a machine learning model that successfully classified among  combinations of any six rainforest fauna sonotypes at a time, as long as each sonotype had at least 3 samples (212 sonotypes). With the techniques of data augmentation and transfer learning, we were able to increase the mean accuracy of our model to 90.4\%, even at extremely small sample sizes (Fig. \ref{6 random all}). Our work advances the field by enabling the classification of \textit{all} sound types with at least 3 labelled instances, including rare ones, those emitted by birds, mammals, amphibians, and insects. With our open-source model, even relatively small projects may be able to use CNNs to classify calls within acoustic biodiversity surveys, focusing on all vocalizing organisms for which at least 3 instances are available. Such capabilities are important in a number of conservation uses, from prioritization surveys when designating protected forest areas \citep{williams2002data}, monitoring the effectiveness of biodiversity conservation projects \citep{burivalova2019works}, or understanding the impact of land use change on biodiversity \citep{powers2019global}.

    
From our results, we conclude that to successfully classify amongst sonotypes represented by a relatively small number of samples each, both transfer learning and data augmentation are necessary (Fig. \ref{fix size under varied sonotype number}). When data augmentation is not used, even with transfer learning,  a training data set that contains $\sim{80}$ labelled occurrences of each sound class (sonotype) is insufficient, yielding only modest accuracy of 75.1\% (Fig. \ref{6 random all}). Yet, even a sample size of 80 can be challenging to achieve for rare species, which might vocalize unpredictably, only a few times a day, or for highly mobile species, which might only rarely pass through the recording site. In cases where obtaining several hundreds or thousands labeled occurrences of each sonotype is not feasible, our results suggest that augmenting the training data by artificially cropping, squeezing, and stretching the sounds, as well as by adding noise, can vastly increase the classification accuracy (Fig. \ref{fix size under varied sonotype number}, \ref{6 random all}). Even at extremely small sample sizes, such as only a handful of examples per sonotype, our model with data augmentation reached high and consistent average accuracy of $\geq{90\%}$. Data augmentation was beneficial regardless of the original training data set being balanced or not, and regardless of the taxonomic group (Fig. \ref{6 random all},\ref{taxonomic group}, Table \ref{tab-our-results}). Interestingly, without data augmentation, increasing the average sample size had a similar effect on accuracy whether the dataset was balanced or not.

Our results add to the emerging evidence on benefits of data augmentation in sound classification. A study of bird call classification found that data augmentation increased the classification accuracy by 9 percentage points, whereby the largest improvements were generated by adding background noise \citep{lasseck2018audio}. Whereas it is not possible to test on our dataset, we hypothesize that the augmentation methods that slightly shift  frequency maybe be useful when classifying calls from different areas, as some species are known to shift the frequency of their song in response to environmental changes or acoustic competition \citep{derryberry2020singing}. Even if one type of augmentation can show larger gain in accuracy than others, for smaller, non-simulated datasets such as ours, it is important to use a variety of augmentation techniques to avoid potential hidden biases. However, data augmentation of spectrograms requires special care, as certain transformations (e.g., flipping or rotating) could result in  spectrograms corresponding to entirely different sound patterns, and that is why we avoided these \citep{nanni2020data}. 

Whereas there is as yet no standardized reporting for the performance of bioacoustic CNN models (but see \citep{knight2017recommendations} for single species classifiers), our results compare favourably with existing literature (Table \ref{tab-models-results} and \ref{litreview}).  Most CNN studies classify sounds from active, directional recordings, such as those available through the Xeno-Canto database (https://www.xeno-canto.org/). When translated to passive soundscapes, the performance typically drops, due in part to the lower quality of signals and to errors introduced with segmentation of complex soundscapes \citep{kahl2021birdnet}. Despite using exclusively passive soundscapes, both for training and testing, our study reaches high levels of accuracy.

\begin{footnotesize}
\begin{table}[htp]
\caption{Accuracy achieved by related studies on animal call classification using neural networks. See  Table \ref{litreview} for model parameters, species numbers, and use of data augmentation. Studies are based on different datasets, therefore the accuracy measures are not directly comparable. mAP = mean average precision; cmAP = class-wide mean average precision; AUC = area under the receiver operating characteristic curve; Rec. = recall (or sensitivity); Spec. = specificity; f1 = f1 Score}.
\label{tab-models-results}

\centering
\begin{tabular}{p{4cm}p{1.5cm}p{1cm}p{1cm}p{1cm}p{1cm}p{1cm}p{1cm}p{1cm}}
  \toprule

Author	&	Accuracy	&	mAP &	cmAP	&	AUC	&	Rec.	&	Spec.	&	f1.	\\

  \midrule
Our model$^1$	&	93.0    94.5	&	94.5    94.5	&	94.5    94.5	&	98.8    98.9	&	93.0	94.5    &	98.2    99.3	&	92.5	94.1    \\
Kahl et al. 2021	&	77.7	&	79.1	&	69.4	&	97.4	&	-	&	-	&	-	\\
LeBien et al. 2020	&	-	&	97.5	&	89.3	&	-	&	-	&	-	&	-	\\
Zhong et al. 2020$^2$	&	-	&	-	&	-	&	97.5    97.9    99.5	&	82.1    84.1    97.7	&	96.9    97.7    96.4	&	-	\\
Tabak et al. 2020$^3$	 &	90	&	-	&	-	&	-	&	-	&	-	&	91	\\
Goeau et al. 2019$^4$	&	83	&	-	&	19.3	&	-	&	-	&	-	&	-	\\
Kahl et al. 2020$^4$	&	-	&	74.5	&	35.6	&	-	&	-	&	-	&	-	\\
Ruff et al. 2021	&	99.5	&	-	&	-	&	-	&	-	&	-	&	-	\\
Ruff et al. 2019$^6$	&	-	&	-	&	0.4-77.1 	&	-	&	63.1-91.5	&	-	&	-	\\
Hidayat et al. 2021	&	97.1	&	95.2	&	97.6	&	-	&	96.4	&	-	&   -   \\
Chen et al. 2020	&	90.2	&	-	&	-	&	94	&	90.9	&	85.3	&	-	\\
Xie et al. 2019	&	86.3	&	92.1	&	-	&	-	&	99.5	&	91.6	&	93.3	\\
Xie and Zhu 2019	&	-	&	-	&	-	&	-	&	-	&	-	&	95.95	\\
Xu et al. 2020$^5$	&	94.7    86.4	&	93.1    86.9	&	-	&	-	&	94.3    85.1	&	-	&	92.9    86.1	\\
Ko et al. 2018	&	97.18	&	-	&	-	&	-	&	-	&	-	&	-	\\
   \bottomrule
\end{tabular}

\begin{tablenotes}
    \item[1] $^1$Sets of 2 numbers represent values for classification with fixed sample sizes (40 samples, experiment 1) per sonotype and with varying sample sizes (averaging 40 to 41 samples, inclusive, experiment 2) per sonotype.
	\item[2] $^2$Sets of 3 numbers represent values before pseudo-labeling, pre-trained with ResNet50, and with pseudo-labeling.
	\item[3] $^3$Pre-print.
	\item[4] $^4$Working paper.
	\item[5] $^5$Sets of 2 numbers represent values for frogs and crickets, respectively.
	\item[6] $^6$Precision, recall, and f1 scores reported for each species separately. Numbers reported represent a range.
\end{tablenotes}
\end{table}
\end{footnotesize}

There are several limitations to our experiments. First, our training, validation, and testing data sets were all from the same set of recordings, obtained at the same set of rainforest sites during the dawn and dusk, which may have resulted in a higher accuracy than if tested at different sites. We invite future studies to test our method by re-training in other forest types across different biogeographic regions, as well as for organisms that vocalize during different parts of the day. Second, vocalizing species are in some cases able to adapt their vocalizations to changing environmental conditions and competing species, including in terms of frequency and the precise nature of the song \citep{derryberry2020singing,grant2010songs}.  Tentatively, we suggest that accuracy would not decrease with shifts in frequency, as such changes are simulated through data augmentation. However, empirical testing to see whether our model is robust under such changes is necessary. Importantly, we only classified amongst six sonotypes during each model run, so as to achieve enough replicates to allow high enough statistical power to reliably test accuracy. Classifying amongst larger number of sonotypes with our model should be further tested.

A potential source of uncertainty is introduced during the manual labeling of soundscapes. Two separately labeled sounds could be classified as one or two sonotypes, depending on the analyst, quality of the recording, and a random error. In our case, we tried to minimize this error by consulting experts in uncertain cases, and revisiting our classification multiple times. This is a time-consuming process and further research should quantify the level of uncertainty due to different analysts. Importantly, the number of sonotypes does not translate to number of species, as some species may produce more than one call type.

Ultimately, for a fully automated classification of biodiversity from soundscapes, three steps are needed: first, the soundscape needs to be segmented in terms of time and frequency, so that individual sounds are isolated. Existing approaches use for example threshold in amplitude or non-negative matrix factorization to detect the beginning and end of a sound in time, alternatively they  cut the soundscape into equal segments of a few seconds each \citep{araya2017warbler,stowell2019automatic,lin2017improving,lin2020source,searfoss2020chipper,kahl2021birdnet}. These approaches work  well in cases where there is little  overlap between individual sounds, but not yet under substantial temporal or frequency overlap, such as is the case in a hyper-diverse rainforest soundscape. Future work is needed to improve on these techniques. Second, the isolated sounds need to be classified as belonging to one of the predefined classes, and our work addresses this step. A final step, which should be addressed by future studies, is to design a model that is able to distinguish previously not encountered sonotypes, and incorporate them into the model.

Conservation projects could eventually use such fully automated model in several ways. First, multiple sites could be surveyed and ranked according to their sonotype richness and turnover across sites, to contribute to prioritization if only some sites can be protected. Species richness has important limitations \citep{hillebrand2018biodiversity} and is only one of many socio-ecological values in conservation prioritization, it nevertheless remains an important conservation metric \citep{game2013six}. Second, repeated soundscape surveys could alert projects to potential loss of certain sonotypes, due to newly emerging threats \citep{sethi2020characterizing}. Third, effectiveness of conservation projects could be evaluated by quantifying changes in sonotypes numbers, such as through Before-After-Control-Intervention or quasi-experimental surveys.

\section{Conclusions}
We have developed an open-source Convolutional Neural Network that can classify amongst common and rare vocalizing mammals, birds, amphibians and invertebrates from passively recorded Borneo's rainforest soundscapes. Using transfer learning and data augmentation, our models achieve $>90\%$ accuracy even at extremely small training data sample size, but neither transfer learning nor data augmentation can reach this level of accuracy alone. Our model is designed to help small to mid-size conservation projects with limited budgets to deploy the automated classification of rainforest soundscapes in biodiversity surveys and evidence-based conservation.

\section{Acknowledgments}

This project was partially funded by The Nature Conservancy and Precious Forests Foundation to Z.B. This work was also partially supported by the Department of Energy [DE-SC0021016 to C.S.L.]. Y.S. was supported by the Holstrom Environmental Research Fellowship. We thank Purnomo, Yuliana Kagan, Arif Rifqi, Tim Boucher, Eddie Game, land owners, and HTCondor advisors at UW-Madison for assistance.

\section{Conflict of Interest}

The authors have no conflict of interest.

\bibliography{references}

\section{Supporting Information}

\begin{footnotesize}
\begin{table}[ht]

\renewcommand{\thetable}{S\arabic{table}}
\setcounter{table}{0} 

\caption{Accuracy measures for the impact of data augmentation and transfer learning when classifying among 2 - 6 sonotypes, averaged across all tested sample sizes. DA = data augmentation; TL = transfer learning; mAP = mean average precision; cmAP = class-wide mean average precision; AUC = area under the receiver operating characteristic curve.}
\label{tab-aug-transfer-compare}

\begin{center}

\begin{tabular}{lrrrrrrrrrr}

 \toprule
& \multicolumn{5}{c}{without DA without TL} & \multicolumn{5}{c}{with DA without TL}\\
\cmidrule(l){2-6} \cmidrule(l){7-11} 
number of sonotypes	&	2 	&	3  &	4 	&	5   &   6 	&	2 	&	3  &	4 	&	5   &   6\\

  \midrule
Accuracy	&   85.7	&   63.9	&   45.9	&   35.1	&   31.0	&   91.2	&   80.7	&   69.8	&   54.1	&   37.7		\\
mAP	&   85.7	&   63.9	&   45.9	&   35.1	&   31.0	&   91.2	&   80.7	&   69.8	&   54.1	&   37.7		\\
cmAP	&   84.9	&   59.8	&   38.0	&   26.4	&   22.1	&   92.0	&   79.6	&   65.8	&   47.1	&   28.2		\\
AUC	&   88.8	&   78.0	&   69.7	&   66.0	&   64.2	&   94.6	&   89.0	&   87.0	&   79.5	&   73.2		\\
Recall	&   85.7	&   63.9	&   45.9	&   35.1	&   31.0	&   91.2	&   80.7	&   69.8	&   54.1	&   37.7	\\
Specificity	&   85.7	&   82.0	&   82.0	&   83.8	&   86.2	&   91.2	&   90.4	&   89.9	&   88.5	&   87.5  	\\
f1 score	&   83.2	&   57.3	&   36.2	&   25.0	&   21.3	&   90.3	&   78.0	&   64.7	&   46.5	&   28.5	\\

\bottomrule
\end{tabular}
\end{center}

\begin{center}

\begin{tabular}{lrrrrrrrrrr}

 \toprule
& \multicolumn{5}{c}{without DA with TL} & \multicolumn{5}{c}{with DA with TL}\\
\cmidrule(l){2-6} \cmidrule(l){7-11} 
number of sonotypes	&	2 	&	3  &	4 	&	5   &   6 	&	2 	&	3  &	4 	&	5   &   6\\

  \midrule
Accuracy	&   84.9	&   69.1	&   64.2	&   54.8	&   57.7	&   93.6	&   90.1	&   92.3	&   91.9	&   91.3		\\
mAP	&   84.9	&   69.1	&    64.2	&   54.8	&   57.7	&   93.6	&   90.1	&   92.3	&   91.9	&   91.3		\\
cmAP	&   83.0	&   64.9	&   58.3	&   47.1	&   49.1	&   94.7	&   91.5	&   93.4	&   92.9	&   92.2		\\
AUC	&   90.2	&   85.2	&   87.1	&   85.6	&   89.0	&   98.1	&   96.2	&   98.0	&   98.5	&   98.6		\\
Recall	&   84.9	&   69.1	&   64.2	&   54.8	&   57.7	&   93.6	&   90.1	&   92.3	&   91.9	&   91.3		\\
Specificity	&   84.9	&   84.6	&   88.1	&   88.7	&   91.5	&   93.6 	&   95.1	&   97.4	&   98.0	&   98.3  	\\
f1 score	&   82.0	&   64.4	&   58.0	&   47.4	&   49.6	&   93.1	&   89.3	&   92.0	&   91.6	&   90.9	\\

\bottomrule
\end{tabular}
\end{center}

\end{table}
\end{footnotesize}

\begin{footnotesize}
\begin{table}[ht]
\renewcommand{\thetable}{S\arabic{table}}

\caption{Accuracy measures for taxonomic groups without and with data augmentation, averaged across all tested sample sizes. mAP = mean average precision; cmAP = class-wide mean average precision; AUC = area under the receiver operating characteristic curve.}
\label{tab-taxonomic-group}

\begin{center}

\begin{tabular}{lrrrrrrrr}

 \toprule
& \multicolumn{2}{c}{Invertebrates} & \multicolumn{2}{c}{Birds}& \multicolumn{2}{c}{Amphibians} & \multicolumn{2}{c}{Mammals}\\
\cmidrule(l){2-3} \cmidrule(l){4-5} \cmidrule(l){6-7} \cmidrule(l){8-9}
Augmentation	&	Without 	&	With  &	Without 	&	With    &	Without 	&	With    &	Without 	&	With \\

  \midrule
Accuracy	&   51.6	&   90.7	&   50.4	&   90.2	&   53.5	&   89.7	&   57.2	&   91.5		\\
mAP	&   58.3	&   90.7	&   56.4	&   90.2	&   57.2	&   89.7	&   61.1	&   91.5		\\
cmAP	&   58.1	&   90.7	&   56.3	&   90.2	&   57.3	&   89.7	&   61.1	&   91.5		\\
AUC	&   82.7	&   97.9	&   82.3	&   97.9	&   85.1	&   98.0	&   87.6	&   98.5		\\
Recall	&   51.6	&   90.7	&   50.3	&   90.2	&   53.5	&   89.7	&   57.2	&   91.5	\\
Specificity	&   90.3	&   98.1	&   90.1	&   98.0	&   90.7	&   97.9	&   91.4	&   98.3	\\
F1 score	&   43.0	&   89.4	&   41.5	&   88.9	&   45.4	&   88.9	&   49.4	&   90.8	\\

\bottomrule
\end{tabular}
\end{center}
\end{table}
\end{footnotesize}

\begin{landscape}
\begin{table}[ht]
\renewcommand{\thetable}{S\arabic{table}}
\caption{Related studies on animal call classification using neural networks. See table 1 for accuracy measures.}
\label{litreview}

\centering
\begin{tabular}{p{3cm}p{5cm}p{1.5cm}p{3cm}p{3cm}p{2cm}p{3cm}}
  \toprule

Author	&	Base model	&	Augment.	&	Input data	&	Sample size	&	Classes	&	Taxonomic groups	\\
  \midrule

Our model	&	VGG19	&	Yes	&	 soundscapes	&	3629	&	448 (6 at a time)	&	Birds, invertebrates, mammals, frogs	\\
  
Kahl et al. 2021	&	ResNet	&	Yes	&	Xeno-canto, Cornell, soundscapes	&	226,078 recordings, max. 500 per class	&	984	&	Birds	\\

LeBien et al. 2020	&	ResNet	&	No	&	Soundscapes	&	86,652 t.p.; 188,908 f.p.	&	24	&	Birds, frogs	\\

Zhong et al. 2020	&	VGG16, ResNet50	&	Pseudo-labeling	&	Soundscapes	&	100,000 p.; 243,000 n.	&	24	&	Birds, frogs	\\

Tabak et al. 2020	&	ResNets	&	No	&	Call library	&	11,514	&	10	&	Bats	\\

Goeau et al. 2019	&	Inception-V3	&	Yes	&	Xeno-canto	&	36,496	&	1500	&	Birds	\\

Kahl et al. 2020	&	Inception, ResNet	&	Yes	&	Xeno-canto	&	50,153	&	659	&	Birds	\\

Ruff et al. 2021	&	6 trainable layers, 4 convolutional layers, 2 fully connected layers	&	Yes	&		&	53,292	&	17	&	Birds	\\

Ruff et al. 2019	&	4 convolutional layers, 2 fully connected layers	&	No	&	Field recordings	&	3000-4000 per class	&	6	&	Birds	\\


Hidayat et al. 2021	&	Adapted from Sprengel et al. 2016	&	Yes	&	Xeno-canto	&	752	&	7	&	Birds	\\

Chen et al. 2020	&	BatNet, 22 convolutional layers	&	No	&	Field recordings	&	130,858	&	36	&	Bats	\\

Xie et al. 2019	&	VGG, SubSpectralNet	&	No	&	Field recordings	&	5428	&	43	&	Birds	\\

Xie and Zhu 2019	&	3 convolutional layers	&	No	&	Xeno-canto	&		&	14	&	Birds	\\

Xu et al. 2020	&	multi-view CNN, 3 views each with 3 convolutional layers	&	No	&		&		&	14; 20	&	Frogs, crickets	\\

Ko et al. 2018	&	5 convolutional layers	& NO	&	soundscapes, ebird.org, Korea Wild Animal Sound Dictionary	&	52,765	&	102	&	Anuran, birds, and insects	\\

\bottomrule
\end{tabular}
\end{table}
\end{landscape}


\begin{figure}[ht]
\renewcommand{\thefigure}{S\arabic{figure}}
\setcounter{figure}{0} 
\centering
\includegraphics[width=14cm]{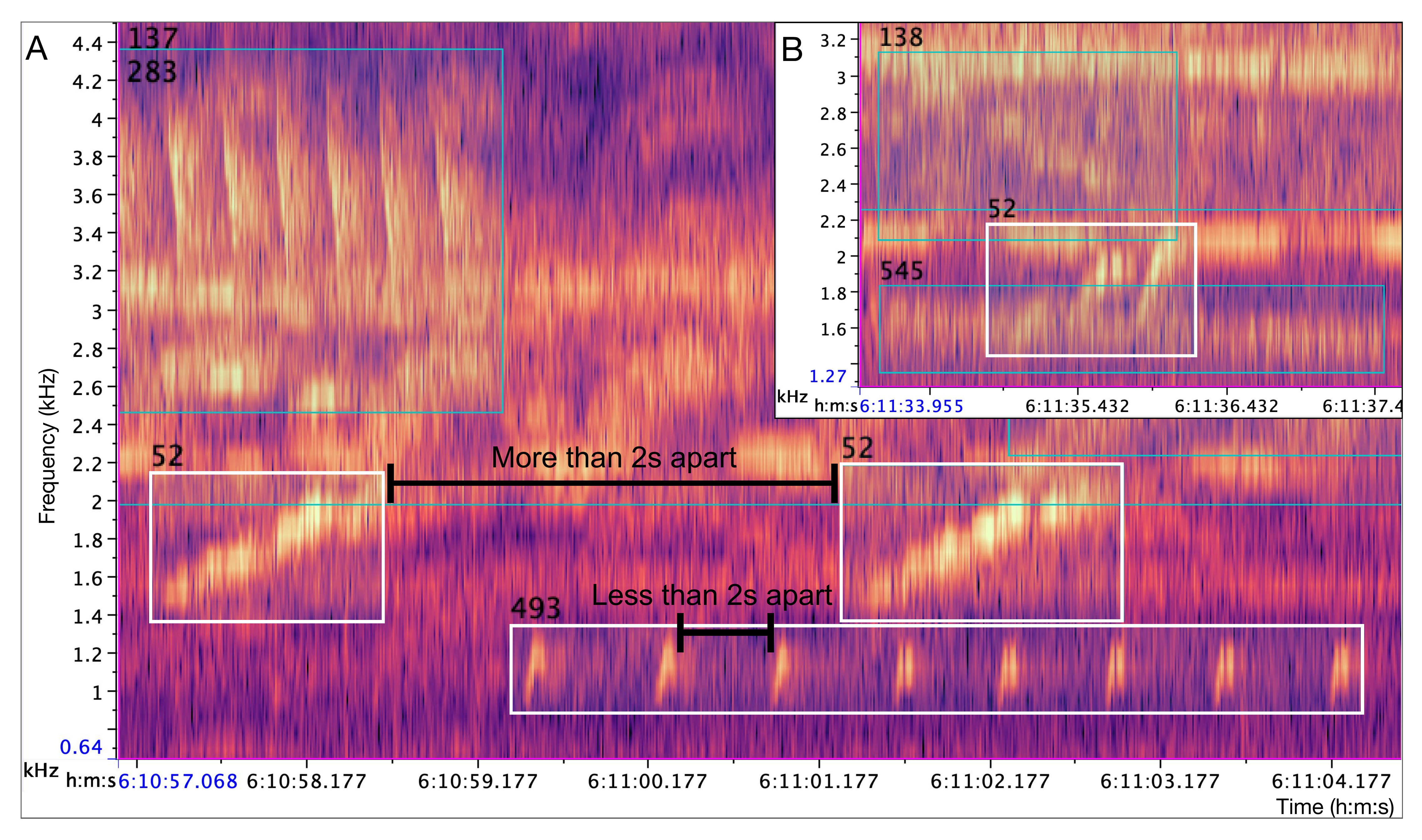}
\caption{A spectrogram illustrating sonotypes 52 and 493, highlighted with white borders (A). Instances of sonotype  52 are more than 2s away from each and so were labeled separately. Instances of sonotype  493 are less than 2s apart and so were labeled together. The same sound is always labeled as the same sonotype (e.g. sonotype  52), even if it occurs at a different location and time of the day (B).}
\label{FigS1}
\end{figure}

\end{document}